# Build Electronic Arabic Lexicon

Nidhal El-Abbadi, Ahmed Khdhair, and Adel Al-Nasrawi
University of Kufa, Iraq

**Abstract:** *There are many known Arabic lexicons organized on different ways, each of them has a different number of Arabic words according to its organization way. This paper has used mathematical relations to count a number of Arabic words, which proofs the number of Arabic words presented by Al Farahidy. The paper also presents new way to build an electronic Arabic lexicon by using a hash function that converts each word (as input) to correspond a unique integer number (as output), these integer numbers will be used as an index to a lexicon entry.*

## 1. Introduction

Al Khalil bin Ahmed Al Farahidy (died in 175 A.H.) had the priority in creating the first Arabic lexicon. As this great scientist was a man of creative mathematical sense which he used in the whole of the specialization fields on which he worked. This great bright thinking scientist spared no effort in his trial for producing the lexicon which could be used for systemizing and restricting language. Al Farahidy used a mathematical method for collecting language pieces and systemizing them, a method through which he managed to get his aim in hand, and no one had the precedence in it.

Al Farahidy noticed that the Arabic language composed of (28) letters all of which reproduced in vocal cords except for (alhamza) (ء) which is produced from glottis; therefore, he found that no words and letter would produce from alhamza, he searched for the Arabic words and organized them in a firm system. According to that order (system) Arabic words were restricted to the syllables of two, three, four, or five letters, no more and no less with one exception for the extra or additional letters that had nothing to deal with the lexical meaning of the base word [5].

This kind of system which is based on restricting the Arabic word between the two and five letters syllable simplified the work of organizing and collecting the vocabularies of language within a precise.

### 1.1. The Steps of Al Farahidy System in Al-Ein Book

Al Khalil invented a new system for organizing dictionary letters relied on their places of articulation, the dictionary was not arranged alphabetically but rather by phonetics, following the pattern of pronunciation of the Arabic alphabet from the deepest letter of the throat (ع)( (ein) to the last letter pronounced by the lips, that being ( ا ) (alph) so he got the following system: zero category is given to (alhamza) since it is produced from glottal place, and not from vocal cords or the mouth. Al Khalil give this system a base for his work to organize his new lexicon, and naming each letter as a book, so he start with al-Ein (ع) lexicon book.

Table 1. Arabic letters according to Al Khalil organization.

| ض | ش | ج | ك | ق | غ | خ | ه | ح | ع |
|---|---|---|---|---|---|---|---|---|---|
| 10 | 9 | 8 | 7 | 6 | 5 | 4 | 3 | 2 | 1 |
| ر | ث | ذ | ظ | د | ت | ط | ز | س | ص |
| 20 | 19 | 18 | 17 | 16 | 15 | 14 | 13 | 12 | 11 |
|   | ء | ا | ي | و | م | ب | ف | ن | ل |
|   | 0 | 28 | 27 | 26 | 25 | 24 | 23 | 22 | 21 |

The second step was to search for all the Arabic words restricted between the two and five syllables, and he collected them according to that. As for the third step in searching for the words and vocabularies within the target language (Arabic), using permute.

For example the sound (ع) can change its location within the two letter syllable two times as it either comes in the first or second position and in the three letter-syllables it will happen three times in the four letter syllable, it is four times while in the five letter syllable it repeats itself five times. Thus, if the second letter with the (ع) in two letter syllable was (م) we would get only two forms (عم, مع). While in the three letters syllable, if the (ع) was combined with (م) and (د) we would have six forms ( عمد, عدم, معد, مدع, دعم, دمع ) and these forms reach the number of (24) in the four letters syllable, while in the five letters syllable they become (120) forms[5]. These were the most important steps taken by Al Farahidy in writing his al-Ein Lexicon.

This method had been criticized due to its difficulty but criticism was lacking as the ones who criticized Al Farahidy method followed his steps just like Al Azhari (died in 370 A.D) in his Lexicon (Al Tahzeeb). He used steps in the very same way of Al Farahidy, as for the other lexicons, they depended heavily on him either in their subject matter or in their methods although they implied some kind of difference.

Based on the method and the way adopted by the writer in writing a lexicon, Dr. Hussein Nassar [5] has classified the Arabic lexicons into several schools.

### 1.2. Lexicon Schools

Dr Hussein Nassar has set the following classification for the Arabic lexicons which falls into four types in accordance with their methods.

#### 1.2.1. The First School

It is also known as al-Ein school in reference to al-Ein book by Al Khalil for its precedence in the Arabic lexicography world and its uniqueness in method which differs from the others in collecting the Arabic vocabularies and this school has the following lexicons:

- Al-Ein book by Al Khalil bin Ahmed Al Farahidy (died in 175 A.H).
- Al Bariaa book by abi Ali Al Qali (died in 356 A.H).
- Al Tahzeeb by abi Mansour Al Azhari (died in 370 A.H).
- Al Muheet by Al Saheb bin Abad (died in 385 A.H).
- Al Muhkam by ibin Sayidah(died in 458 A.H).

#### 1.2.2. The Second School

It is known Al Jamhara, and it followed permute system set by Al Farahidy in his book Al Ein in addition to its pursuing of the alphabetical order in organizing the words and it has the following lexicons:

- Al Jamhara book by abi Bakr Mohammed bin Duraid (died in 321 A.H).
- Al Maqayis book by Ahmed bin Faris (died in 395 A.H).
- Al Mujmal book by Ahmed bin Faris.

#### 1.2.3. The Third School

It is known as Al Sihah School. This one relied on using (section and chapter) system in which the last letter is a chapter and the first letter is a section, and it has the following lexicons:

- Tajulughah and Sihah Al Arabia by abi Nasar Ismail bin Hammed Al Jauhri( died about 400 A.H).
- Al Abab Al Zakhir Wal Lubab Al Fakhir by Al Hassan bin M. Al Sagani (died in 650 A.H).
- Lisan Al Arab by ibin Mandour (died in 711 A.H).
- Al Qamos Al Muheet by Al Fairouz Abadi( died in 817 A.H).
- Al Tiraz Alau'l wal Al Kanaz Fema Alaih Lught Al Arab Al Mua'l by ibin Ma'ssoum Al Madani(died in 1120 A.H)
- Tajullarous by M. bin Murtada Al Zubaidi(died in 1205 A.H).

#### 1.2.4. The Fourth School

It is called Al Assas School. It used the widely used and easiest method of the alphabetical order in which words are arranged from the first letter to the last one in the word according to the main letters excluding the extra ones, these are:

- Assas Al Balaghah by Jar Allah Mahmoud Al Zamkhashari(died in 438 A.H).
- Al Misbah Al Munir by Al Fayoumi (died in 760 A.H).
- The whole of written lexicons in the modern age such as Muheet Al Muheet by Butras Al Bustani, Aqrab Al Mawared fi Fasih Al Arabia and Al Shawared by Saed Al Khouri Al Sharnoubi and the other modern lexicons.

## 2. Hash Table

A hash table works by transforming the key using a hash function into a *hash*, a number that is used as an index in an array to locate the desired location ("bucket") where the values should be. The number is normally converted into the index by taking a module, or sometimes bit masking is used where the array size is a power of two. The optimal hash function for any given use of a hash table can vary widely, however, depending on the nature of the key.

Typical operations on a hash table include insertion, deletion and lookup (although some hash tables are precalculated so that no insertions or deletions, only lookups are done on a live system). These operations are all performed in amortized constant time, which makes maintaining and accessing a huge hash table very efficient.

- Perfect Hashing: if all of the keys that will be used are known ahead of time, and there are no more keys that can fit the hash table, perfect hashing can be used to create a perfect hash table, in which there will be no collisions. If minimal perfect hashing is used, every location in the hash table can be used as well. Perfect hashing gives a hash table where the time to make a lookup is constant in the worst case. This is in contrast to chaining and open addressing methods, where the time for lookup is low on average, but may be arbitrarily large. There exist methods for maintaining a perfect hash function under insertions of keys, known as dynamic perfect hashing. A simpler alternative, that also gives worst case constant lookup time, is cuckoo hashing.
- Problems with Hash Tables: although hash table lookups use constant time on average, the time spent can be significant. Evaluating a good hash function can be a slow operation. In particular, if simple array indexing can be used instead, this is usually faster.

Hash tables in general exhibit poor locality of reference-that is, the data to be accessed is distributed seemingly at random in memory. Because hash tables cause access patterns that jump around, this can trigger microprocessor cache misses that cause long delays. Compact data structures such as arrays, searched with linear search, may be faster if the table is relatively small and keys are cheap to compare, such as with simple integer keys. According to Moore's Law, cache sizes are growing exponentially and so what is considered "small" may be increasing. The optimal performance point varies from system to system.

More significantly, hash tables are more difficult and error-prone to write and use. Hash tables require the design of an effective hash function for each key type, which in some situations is more difficult and time-consuming to design and debug than the simple comparison function required for a self-balancing binary search tree. In open-addressed hash tables it is fairly easy to create a poor hash function.

## 3. Propose Algorithm

Due to advances in computer technology, language dictionaries are gradually gaining additional importance. Great achievements have been reported in areas such as natural language processing, speech recognition and other AI applications. Most of these applications require the availability of a dictionary that can be maintained and accessed in a very efficient way. Dictionaries are also required in conventional applications such as spelling correction and data base management. Various types of Arabic's search engines are significantly impaired because of the inability to find character-to-character correspondence between search terms and variant match items.

In this paper, we present new way to build an electronic Arabic lexicon depending on building a hash table using a word as a key to produce a corresponding integer index, this can be accomplished by using one to one and onto a mathematical relation. To build a lexicon we will be restricted with the following Al Farahidy idea:

1. Galal Al Den Al Suoty [4] ( died in 911 A.H ) says that Al Khalil bin Ahmed Al Farahidy counted the root of Arabic words (the used and unused words) with the four syllable (two , three, four, and five) without repeat equal to (12305412) words and vocabularies, without declaring how to count them.

To prove this theory we build mathematical relation which helps to count the number of roots of the Arabic language, the relation is: number of words for syllable = N! / (N – R)!... ( 1 ).
where
N: is the number of Arabic letters and equal to (28)
R: the syllable (R = {2, 3, 4, 5})
Number of words for R=2, is = 28! / (28 – 2)! = 756
Number of words for R=3, is = 28! / (28–3)! =9650
Number of words for R=4, is = 28! / (28–4)! = 490400
Number of words for R=5, is = 28! / (28–5)! = 11793600
Total number of words for the four syllables = 12305412 words and vocabularies. This is the same number for Al-Farahidy. The second thing is using Table 1 as a letters weight.

Table 2. Relation between words letter and its weight (according to Table(1)) and corresponding indexes. Where (*L.W* mean letter weights, and *W.L* mean word letters).

| L.W   | 1, 1        | 2, 1        | ... | 28, 1         |
| ----- | ----------- | ----------- | --- | ------------- |
| W.L   | ع, ع        | ح, ع        |     | ا, ع          |
| Index | 1           | 2           |     | 28            |
| L.W   | 1, 2        | 2, 2        | ... | 28, 2         |
| W.L   | ع, ح        | ح, ح        |     | ا, ح          |
| Index | 29          | 30          |     | 56            |
| L.W   | 1, 28       | 2, 28       | ... | 28,28         |
| W.L   | ع, ا        | ح, ا        |     | ا, ا          |
| Index | 757         | 758         |     | 784           |
| L.W   | 1, 1, 1     | 2, 1, 1     | ... | 28, 1, 1      |
| W.L   | ع, ع, ع     | ح, ع, ع     |     | ا, ع, ع       |
| Index | 785         | 786         |     | 812           |
| L.W   | 1, 2, 1     | 2, 2, 1     | ... | 28, 2, 1      |
| W.L   | ع, ح, ع     | ح, ح, ع     |     | ا, ح, ع       |
| Index | 813         | 814         |     | 840           |
| L.W   | 1, 1, 1, 1, 1 | 2, 1, 1, 1, 1 | ... | 28, 1, 1, 1, 1 |
| W.L   | ع, ع, ع, ع, ع | ح, ع, ع, ع, ع |     | ا, ع, ع, ع, ع |
| Index | 637393      | 637394      |     | 637420        |

2. According to previous restriction each letter in word will be replaced with corresponding number (weight) in Table 1, and then according to the following relation each set of numbers can produce one unique number (we used Table 3 below as a base for derivative relation 2): $d_5 \cdot (28)^4 + d_4 \cdot (28)^3 + d_3 \cdot (28)^2 + (d_2 - 1) \cdot 28 + d_1$... (2)
3. Relation 2 in step 3 is (one to one , and onto) and it used to convert each word to corresponding unique

index, each index represents one word of lexicon according to Al Farahidy definition.
4. To search the lexicon for a specific word meaning we convert a word to its corresponding index according to Table 1 and relation 2 and then direct access to the meaning of word in counted index.
5. Table 3 shows some of index for different words.

Table 3. Lexicon indexes for different words.

| Word | Syllable | Letters ($d_5,d_4,d_3,d_2,d_1$) | Corresponding Numbers According to Table 1 ($d_1,d_2,d_3,d_4,d_5$) | Index According to Equation 2 |
|---|---|---|---|---|
| عم | 2 | ع, م | 1, 25 | 673 |
| قد | 2 | ق, د | 6, 16 | 426 |
| عمر | 3 | ع, م, ر | 1, 25, 20 | 16353 |
| كتب | 3 | ك, ت, ب | 7, 15, 22 | 17647 |
| جواد | 4 | ج, و, ا, د | 8, 26, 28, 16 | 373892 |
| دحرج | 4 | د, ح, ر, ج | 16, 2, 20, 8 | 191340 |
| سفرجل | 5 | س, ف, ر, ج, ل | 12, 23, 20, 8, 21 | 13099700 |
| أقشعر | 5 | ا, ق, ش, ع, ر | 28, 6, 9, 1, 20 | 12322296 |

## 4. Conclusions

This paper presents a mathematical method to count a number of words which proved Al Farahidy theory. Our technique provides direct access to any word in lexicon (by counting and matching index) instead of sequential search for word letters, one letter after another. This way can be applied to any dictionary with any language. By using relation 2, each word produces unique number, in which there will be no collisions, which means every location in the hash table can be used. The proposed hash function dose not need to take a module, or bit masking to convert number into the index. Typical operations on a hash table include insertion; deletion and lookup are done on a live system. The proposed algorithm used arrays with integer keys or indexes, and then the searched time (linear search) may be faster. Although convert words (letters) to corresponding integers they use constant time on average, but the time spent can be significant. By using this relation we get more than (12305412**)** words, this due to return the words of repeated letters to its origin in the (2, 3, 4, 5) syllables like (الصلصله, والزلزله) which are classified as two syllables (صل, زل) [5], also relation return not meaning words which can be delete from lexicon to reduce its size, like (عع, عح, بغبغ).

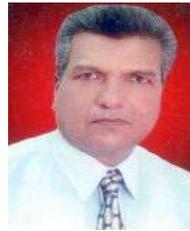
**Nidhal El-Abbadi** received two BSc, in chemical engineering from Baghdad University, and in computer science from al Mustansiryiaa University, received his MSc and PhD in computer science from Informatics Institute for postgraduate studies, his research interests are in image processing, pattern recognition, and steganography.

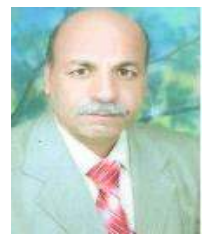
**Adel Al-Nasrawi** received BSc in civilian engineering from Kufa University, BSc and MSc in Arabic language science from Kufa University, currently, he is PhD student in Arabic language Department, Kufa University, Iraq.

.